\documentclass[conference]{IEEEtran}
\IEEEoverridecommandlockouts
\usepackage{cite}
\usepackage{amsmath,amsfonts}
\usepackage{graphicx}
\usepackage{multirow}
\usepackage{booktabs}
\usepackage{textcomp}
\usepackage{url}
\usepackage{needspace}
\usepackage{flushend}
\usepackage{xcolor}
\title{PEETS: Preference-aware Embeddings with Explicit Tree Structure for Auxiliary Product Bundling}

\author{
\IEEEauthorblockN{Maolei Huang\textsuperscript{1,*}, Jing Tian\textsuperscript{2}, Dexuan Duan\textsuperscript{2}, Yan Chen\textsuperscript{2}, Huawei Cao\textsuperscript{2}, Shuhan Song\textsuperscript{2,*,\textdagger}}
\IEEEauthorblockA{\textsuperscript{1}Fliggy, Alibaba Group, Beijing, China\\
\textsuperscript{2}University of Chinese Academy of Sciences, Beijing, China}
\IEEEauthorblockA{huangmaolei.hml@gmail.com, tianjing923.tj@gmail.com, ddx20010123@gmail.com\\
chenyan@gmail.com, caohuawei@ict.ac.cn, songshuhan@ict.ac.cn}
\thanks{\textsuperscript{*}Maolei Huang and Shuhan Song contributed equally.
\textsuperscript{\textdagger}Corresponding author: Shuhan Song.}
}

\newcommand{\figdesc}[1]{}
\newcommand{\papertablestretch}{1.00}

\begin{document}
\maketitle
\begin{abstract}
Auxiliary products such as flight protection and flexible-change services are recommended alongside primary travel purchases. This setting creates a granularity ambiguity: SKUs within one product type share semantics but differ in price and coverage, while rejecting one offer does not imply disinterest in the entire type. We present \textbf{PEETS} (\textbf{P}reference-aware \textbf{E}mbeddings with \textbf{E}xplicit \textbf{T}ree \textbf{S}tructure), a training-time framework for auxiliary product bundling. PEETS regularizes an existing click--conversion predictor through two conditional comparisons. Hierarchical Contrastive Loss uses the pair's common ancestor as a reference; Negative Contrastive Loss conditions structural sampling and comparison shifts on logged rejection events. Both act on shared SKU representations during training. This separates catalog relatedness from individual preference supervision and leaves the predictive serving path intact. PEETS achieves 78.20\% AUC, exceeding TWIN V2 by 0.90 percentage points and the matched base model by 1.15 points. In a 14-day online evaluation, gross auxiliary GMV per exposed user-day improves by 8.20\% over Policies, 3.06\% over Base, and 0.84\% over TWIN V2, with all three user-bootstrap 95\% intervals above zero. Bundling rate is 0.51 points below Policies. The cancellation interval's upper limit exceeds the prespecified margin. PEETS adds no auxiliary-loss computation to serving and records 42.1~ms p95 latency, compared with 42.0~ms for Base.
\end{abstract}

\section{Introduction}
\begin{figure}[!ht]
\centering
\includegraphics[width=\linewidth]{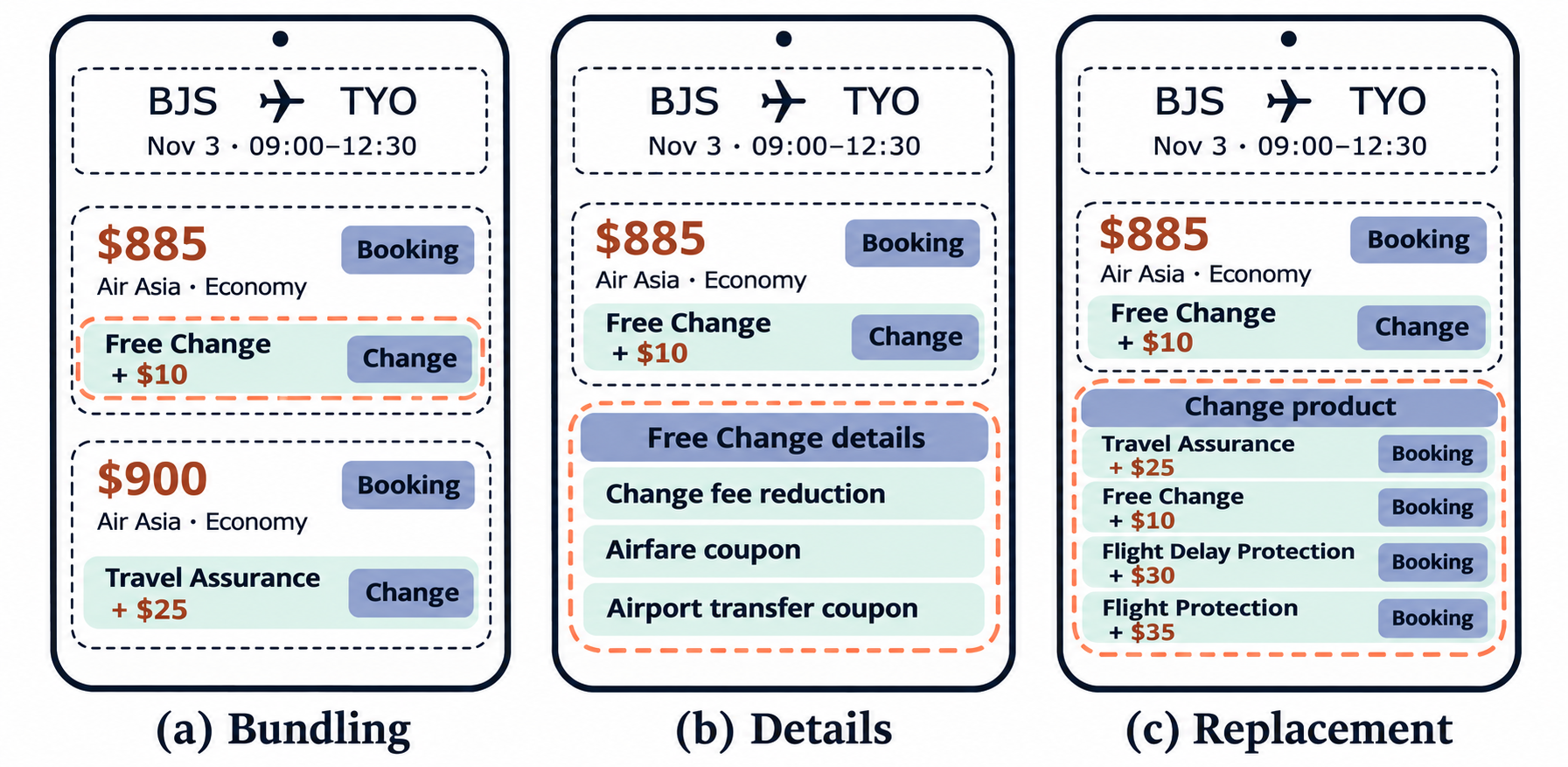}
\caption{Schematic auxiliary-product interfaces: (a) bundling, (b) details, and (c) replacement offers.}
\label{fig:aux1}
\figdesc{Three aligned phone screens show bundled flight offers, benefits of the selected add-on, and alternative auxiliary products. Warm-colored dashed outlines identify the relevant selection, details, and replacement areas.}
\end{figure}

Travel purchases involve choosing an itinerary, an auxiliary product type, and an offer with specific price and coverage. Ranking must distinguish catalog-related offers that serve different needs. Figure~\ref{fig:aux1} shows the selection and replacement interfaces. Here, auxiliary product bundling ranks eligible add-ons for a selected itinerary; joint flight--add-on optimization is outside our scope.

We call this difficulty \emph{granularity ambiguity}. An auxiliary product type, or \emph{xitem}, groups related stock keeping units (SKUs), but its SKUs are not interchangeable: rejecting limited coverage can coexist with interest in a comprehensive offer. Representations must preserve shared semantics and decision-relevant distinctions, while distinguishing explicit rejection from an unchosen offer.

PEETS uses catalog structure to condition training comparisons while retaining clicks and purchases as preference supervision. Hierarchical Contrastive Loss (HCL) references a pair's type-specific common ancestor; Negative Contrastive Loss (NCL) conditions comparisons on logged rejections and sampled reference buckets. Both update shared SKU representations without changing the predictive serving path.

We specify these similarities, sampling rules, and gradient paths, evaluate matched-backbone and NCL component controls, and assess hierarchy sensitivity and online commercial outcomes.

\section{Related Work}
\subsection{Recommendation and Multi-task Learning}
Feature interaction and behavior modeling are central to industrial recommendation. Wide \& Deep~\cite{cheng2016wide} combines linear and nonlinear features; DIN~\cite{zhou2018deep} and DIEN~\cite{zhou2019deep} model target-related interest and its evolution. SIM~\cite{pi2020search}, TWIN~\cite{twin2023}, and TWIN V2~\cite{twinv2024} address long-history retrieval and interest extraction. SRP4CTR~\cite{srp4ctr2024} transfers sequential pre-training into CTR prediction. Flight-itinerary ranking~\cite{huang2020personalized} and nearby-flight recommendation~\cite{huang2025context} provide related travel applications, with different candidate spaces and prediction targets.

ESMM~\cite{ma2018entire} models click--conversion dependencies, while MMoE~\cite{ma2018modeling} and PLE~\cite{tang2020progressive} control information sharing across tasks. AITM~\cite{xi2021modeling}, HoME~\cite{wang2025home}, and DTN~\cite{bi2024dtn} further study task dependencies and interactions. These architectures are relevant backbones or comparators. A comparison with them alone, however, cannot isolate the effect of the tree or the contrastive objectives; matched-backbone ablations are necessary.

\subsection{Hierarchical and Contrastive Representations}
Contrastive learning shapes representations through positive and negative comparisons~\cite{chen2020simple,wang2020understanding}. Supervised and label-aware variants exploit additional relational information~\cite{khosla2020supervised,pan2022improved,liu2022label}. Hierarchical methods include multi-label penalties~\cite{zhang2022use}, hierarchical prototypes~\cite{guo2022hcsc}, and category-aware query classification~\cite{zhu2023hcl4qc}. HAKG~\cite{hakg2022} incorporates hierarchical knowledge into recommendation. Related work also studies cross-domain matching~\cite{chang2025separated}, positive selection in sequential recommendation~\cite{wang2024relative}, and inventory-add prediction at Fliggy~\cite{li2025contrastive}.

PEETS places hierarchical references inside the contrastive similarity: HCL conditions on the SKU pair, and NCL on a rejection event. We therefore compare established recommenders with backbone-matched controls receiving the same catalog information and ordinary contrastive supervision.

\section{Problem Formulation}\label{sec:problem}
\subsection{Decision Scope and Feedback}
Let $u\in\mathcal U$ denote a user and $q=(o,d,t)$ an origin--destination--departure-time query. The information available before a recommendation consists of the query, historical behavior $\mathcal H_u$, and context $\mathcal C$. Let $\mathcal I(q)$ be the eligible auxiliary SKUs for the selected itinerary. Each $i\in\mathcal I(q)$ belongs to an xitem $c(i)$ and has catalog attributes, including price and coverage. Eligibility and availability are determined before ranking.

A training record distinguishes offer exposure, an xitem click, and a SKU purchase, with labels $y_c^{\mathrm{click}}$ and $y_i^{\mathrm{buy}}$. Explicit rejection is a separately logged action, such as replacing or removing a displayed offer. Non-purchase alone is insufficient. The event definitions and attribution windows appear in Section~\ref{sec:data}; they define which interactions can supervise prediction and which can trigger rejection-conditioned regularization.

The hierarchy is a rooted tree with SKU leaves and attribute-bucket internal nodes. Shared ancestors encode relatedness, not identical user preference: rejecting one SKU does not label every descendant of its ancestor as undesirable.

\subsection{Prediction and Commercial Ranking}
For the fixed itinerary, the commercial ranking objective combines relevance and price. Let $p_i^{\mathrm{click}}$ be the probability of clicking the relevant product type and $p_i^{\mathrm{buy}\mid\mathrm{click}}$ the conditional probability of buying SKU $i$. If purchase follows that click event, an expected gross transaction-value score is
\begin{equation}
\label{eq:GMV}
 \mathrm{pGMV}_i=p_i^{\mathrm{click}}p_i^{\mathrm{buy}\mid\mathrm{click}}\mathrm{Price}_i.
\end{equation}
This quantity is distinct from net platform revenue, which can depend on commissions, refunds, and settlement rules. The revenue definition used in the online evaluation is specified in Section~\ref{sec:online}.

Bundling rate measures the number of auxiliary units purchased per primary ticket purchased within the same observation window:
\begin{equation}
\label{eq:br}
 \mathrm{BR}=\frac{\text{auxiliary units purchased}}{\text{primary tickets purchased}}.
\end{equation}
When multiple auxiliary units can accompany one ticket, this is a unit ratio rather than the fraction of tickets with at least one attachment. The same denominator is used across experiment arms.

A price-weighted serving rule then trades off purchase probability and transaction value. We write the log-price surrogate as
\begin{equation}
\label{eq:score}
 \mathrm{Score}_i=p_{c(i)}^{\mathrm{click}}p_i^{\mathrm{buy}\mid\mathrm{click}}
 [\log(\mathrm{Price}_i/\mathrm{CNY})]^{\alpha},
\end{equation}
where $c(i)$ is the xitem containing SKU $i$ and $\alpha\geq0$. Prices are expressed in CNY, and the logarithm takes the dimensionless ratio $\mathrm{Price}_i/\mathrm{CNY}>1$. Equation~\ref{eq:score} is a ranking surrogate, not Equation~\ref{eq:GMV}. Model comparisons hold its configuration fixed, while a separate experiment studies $\alpha$.

\begin{figure*}[!t]
\centering
\includegraphics[width=\textwidth,height=0.55555556\textwidth]{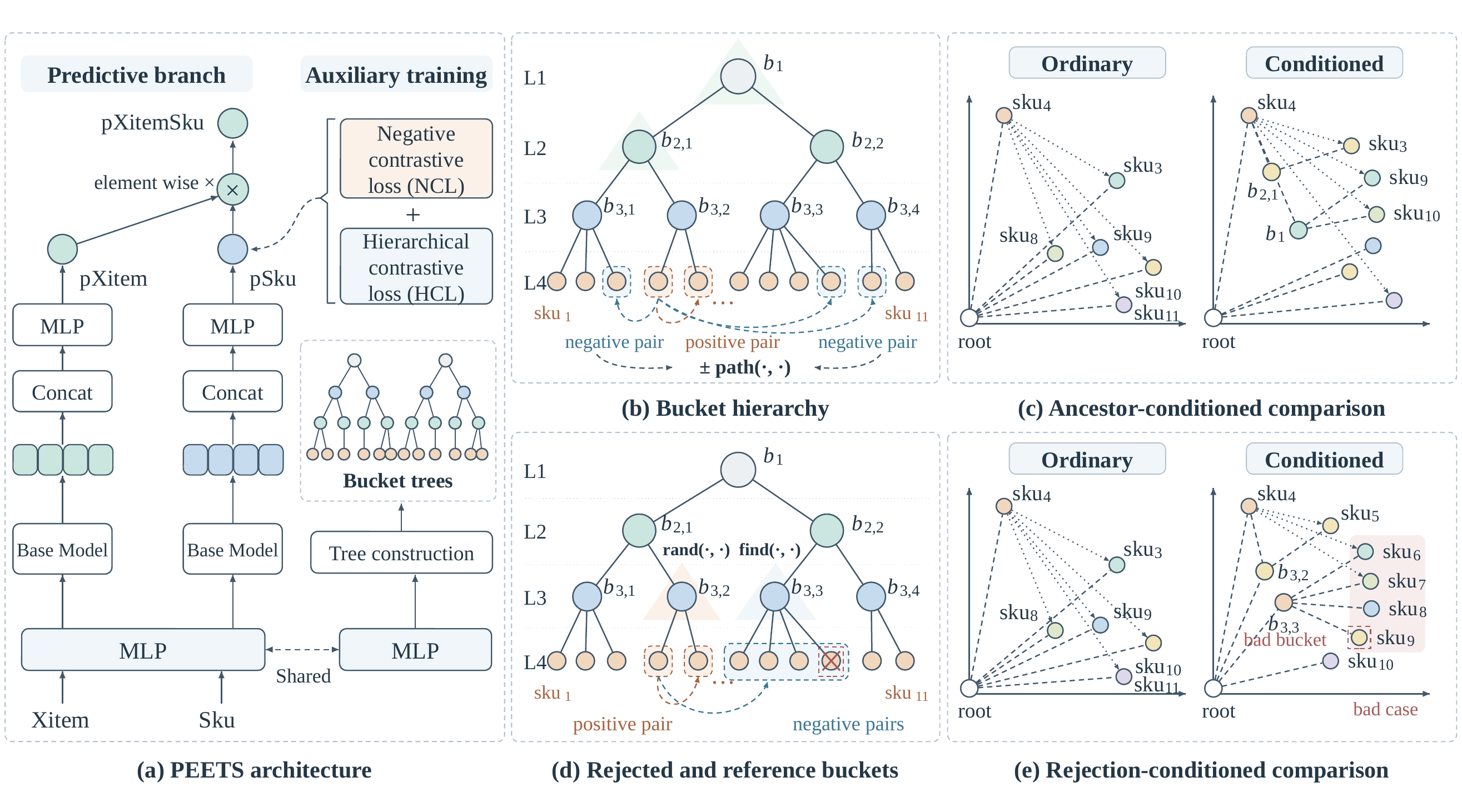}
\caption{PEETS framework. (a) Predictor and training-only auxiliary branch; dashed links denote sharing and training dependencies. (b) Simplified hierarchy (five levels in experiments). (c) Ancestor-conditioned comparisons. (d) Rejected and reference buckets; structural pairs are not preference labels. (e) Rejection-conditioned comparisons; ``bad bucket'' is event-specific. In (c,e), ordinary comparisons are on the left. Here $b$ denotes a bucket and L1 the root. Geometries are schematic, not measured projections.}
\label{fig:PEETS}
\figdesc{A five-panel white-background composite: the PEETS predictor and auxiliary training branch; a four-level bucket hierarchy; ordinary and ancestor-conditioned geometry; rejected and reference buckets; and ordinary and rejection-conditioned geometry.}
\end{figure*}

\section{PEETS}
\subsection{Overview}
Figure~\ref{fig:PEETS}(a) shows the predictive and auxiliary branches. The predictive branch contains xitem-click and SKU-conversion heads. The auxiliary branch compares shared SKU representations under hierarchical and rejection-conditioned transformations. Bucket lookup and contrastive sampling are training operations; online scoring uses the predictive heads and the configured ranking rule.

\subsection{Candidate-aware Prediction}
The inputs contain five groups: user profile, itinerary query, historical behavior, candidate product, and context. Let their embeddings be $e_u,e_q,e_{\mathcal H_u},e_a,e_{\mathcal C}$, where candidate $a$ denotes either an xitem or a SKU. The candidate-specific input is
\begin{equation}
\label{eq:input}
 \mathbf x_a=e_u\oplus e_q\oplus e_{\mathcal H_u}\oplus e_a\oplus e_{\mathcal C}.
\end{equation}
Target-attention backbones also condition the history encoder on $a$. The embeddings follow standard dense representation learning~\cite{mikolov2013distributed}; the experimental backbone and shared layer are a shared MLP with 256/128 hidden units and the 128-dimensional SKU embedding table.

For task $t\in\{\mathrm{xitem},\mathrm{sku}\}$, let $h_t$ denote its encoder and $\sigma$ the sigmoid function. The prediction is
\begin{equation}
\label{eq:model}
 \hat y_{t,a}=\sigma\!\left(w_t^\top h_t(\mathbf x_a)+b_t\right).
\end{equation}
The figure's pXitemSku is the product of these outputs. Auxiliary gradients reach the candidate path through the shared SKU representation in Equation~\ref{eq:input}; no auxiliary features are added at serving time.

Let $\mathcal S_t$ be the eligible labeled records for task $t$. The normalized binary cross-entropy is
\begin{equation}
\label{eq:bce}
 \mathcal L_t=-\frac{1}{|\mathcal S_t|}\sum_{n\in\mathcal S_t}
 \bigl[y_n\log\hat y_n+(1-y_n)\log(1-\hat y_n)\bigr].
\end{equation}
The conditional conversion head is trained on exposed records with an xitem click, without sampling or propensity weighting. A clicked-but-unpurchased record is positive for the click task and negative for conditional conversion; the two tasks therefore use separate labels and eligibility masks.

\subsection{Reproducible Bucket Hierarchy}

Figure~\ref{fig:PEETS}(b) shows the rooted tree $\mathcal T=(\mathcal V,\mathcal E)$: SKU leaves and internal type/attribute nodes, each non-root node having one parent. The root is level one. The SKU-to-xitem mapping, attribute order, interval boundaries, and parent map determine assignment and ancestor lookup.

The experimental hierarchy uses training-catalog price quantiles, attribute order xitem, coarse price, fine price, SKU, and maximum level 5. Its reference number of leaf-level buckets is 15, counted per xitem. Boundaries are estimated from the training-period catalog; duplicate values and interval endpoints follow merged duplicate cut points and right-closed intervals. 

Adjacent fine price intervals merge into coarser intervals; categories follow the catalog taxonomy. Each bucket has an independently trainable embedding, distinct from its descendant SKU vectors. HCL uses the bucket parameter; NCL separately computes the rejected bucket's SKU mean.

The bucket count is selected using only the validation set; the test set is reserved for final evaluation. Tree boundaries use the training catalog only, and each checkpoint records its tree version. 

\subsection{Hierarchical Contrastive Loss}
For a SKU anchor $i$, let $p(i)\ne i$ be a structural positive drawn from the designated bucket. Let $\mathcal N_i$ contain distinct negative candidates excluding $i$ and $p(i)$, and let $\mathcal A_i=\{p(i)\}\cup\mathcal N_i$. With a similarity function $s$ and temperature $\tau>0$, define
\begin{equation}
\label{eq:cl}
 \ell_i(s)=-\log\frac{\exp(s(e_i,e_{p(i)})/\tau)}
 {\sum_{j\in\mathcal A_i}\exp(s(e_i,e_j)/\tau)}.
\end{equation}
The positive is sampled uniformly within the finest bucket and included in the denominator. Negative candidates comprise 128 distinct in-batch SKUs after excluding the anchor and positive; singleton anchors are skipped. Same-bucket membership expresses structural relatedness, not an observed click or purchase label.

Let $b^-(i,j)$ be the common xitem ancestor and $b^+(i,j)$ the deepest common price-bucket ancestor. If the required ancestor type is absent, that component uses the root embedding. These type-specific references belong to the same configured tree. With $\varepsilon>0$ for numerical stability, write
\begin{equation}
\label{eq:cos}
 \operatorname{cos}_{\varepsilon}(a,b)=
 \frac{a^\top b}{\max(\|a\|_2\|b\|_2,\varepsilon)}.
\end{equation}
The two transformed similarities are
\begin{equation}
\label{eq:hcl-sim}
 s^{\pm}(e_i,e_j)=\operatorname{cos}_{\varepsilon}
 \bigl(e_i\pm e_{b^\pm(i,j)},e_j\pm e_{b^\pm(i,j)}\bigr).
\end{equation}
The subtraction form re-centers embeddings around shared category information; the addition form emphasizes a common attribute reference. As illustrated in Figure~\ref{fig:PEETS}(c), the transformation depends on the pair's structural relationship rather than treating all comparisons identically.

For a valid anchor set $\mathcal B$, each HCL component is
\begin{equation}
\label{eq:hcl}
 \mathcal L^{\pm}=\frac{1}{|\mathcal B|}\sum_{i\in\mathcal B}\ell_i(s^{\pm}).
\end{equation}
The xitem component uses subtraction and the price component addition, with $\mathcal L_{\mathrm H}=\mathcal L^-+\mathcal L^+$. Each component uses its type-specific reference above; an empty valid-anchor set contributes zero.

For either component's selected reference $e_b$, write $e_i=e_b+r_i$ and $e_j=e_b+r_j$. The transformed comparison is then between residuals $r_i$ and $r_j$; addition instead compares $2e_b+r_i$ with $2e_b+r_j$. When ancestors differ across pairs, a single global embedding bias cannot generally absorb the transformation. These are origin-dependent angular comparisons, not projections or guarantees of semantic disentanglement. The sign choice is an inductive bias whose value is assessed against ordinary contrastive learning and bucket-feature controls.

\subsection{Rejection-conditioned Contrastive Loss}
An explicit rejection belongs to a user decision, not permanently to a product. Write an observed rejection as $r=(u,q,k_r^-)$, where $k_r^-$ is the rejected displayed SKU. A subsequently selected replacement is recorded separately.

Define $b_r^-=\operatorname{parent}(k_r^-)$ and sample a reference bucket $b_r^{\mathrm{ref}}$ from
\begin{equation}
\label{eq:reference}
 \{b:\operatorname{level}(b)=\operatorname{level}(b_r^-),\;
 k_r^-\notin\operatorname{leaves}(b)\}.
\end{equation}
The reference bucket is sampled uniformly at the same level, excluding the rejected bucket. A sampled bucket is a structural reference, not evidence that its unexposed offers are preferred by the user.

Let $\mathcal P_r$ contain the valid anchor--positive pairs for event $r$. Anchors are sampled uniformly within the reference bucket, each with one same-bucket positive and 128 negative SKUs. The anchor $i$ is distinct from the rejected SKU $k_r^-$; in Figure~\ref{fig:PEETS}(d), the pair lies in the reference bucket.

We write the rejected-bucket aggregate and the shift as
\begin{equation}
\label{eq:shift}
\begin{aligned}
 e_{\mathrm{neg}}(r)&=\operatorname{sg}\!\left(
 \frac{\sum_{k\in\operatorname{leaves}(b_r^-)}e_k}
 {|\operatorname{leaves}(b_r^-)|}\right),\\
 \Delta_r&=e_{b_r^-}-e_{b_r^{\mathrm{ref}}}-e_{\mathrm{neg}}(r).
\end{aligned}
\end{equation}
Here $\operatorname{sg}$ denotes stop-gradient: the SKU mean is recomputed each batch, but this aggregate does not backpropagate into its descendant vectors. For any comparison attached to the same event, the similarity is
\begin{equation}
\label{eq:ncl-sim}
 s_r'(e_i,e_j)=\operatorname{cos}_{\varepsilon}(e_i+\Delta_r,e_j+\Delta_r).
\end{equation}
Let $\ell_{r,i}$ denote Equation~\ref{eq:cl} with the positive and candidate set for event $r$. The event-normalized objective is
\begin{equation}
\label{eq:ncl}
 \mathcal L_{\mathrm N}=\frac{1}{|\mathcal R|}\sum_{r\in\mathcal R}
 \frac{1}{|\mathcal P_r|}\sum_{(i,p)\in\mathcal P_r}\ell_{r,i}(s_r').
\end{equation}
Here $\mathcal R$ contains eligible rejection events with valid pairs. If a batch has no such event, its NCL contribution is defined as zero. The implementation averages anchors within each event, then averages across events.

Figure~\ref{fig:PEETS}(e) illustrates the rejection-conditioned comparison.

\textbf{Gradient paths.} The trainable $e_{b_r^-}$ and detached $e_{\mathrm{neg}}(r)$ do not algebraically cancel. Shifted similarities update compared SKU vectors and bucket parameters; only the aggregate is detached, not every use of its descendant SKU vectors.

\textbf{Geometry and conditioning.} NCL changes comparison angles, not the pair's Euclidean difference: $\|(e_i+\Delta_r)-(e_j+\Delta_r)\|_2=\|e_i-e_j\|_2$. NCL conditions shared representation learning on rejection-bucket statistics; user- and query-specific prediction remains in the supervised branch. False-negative correction~\cite{chuang2020debiased} is not part of this structural sampler.

\subsection{Joint Objective and Training}
With the overall scale fixed to one, the joint objective is
\begin{equation}
\label{eq:objective}
 \mathcal L=\mathcal L_{\mathrm{xitem}}+r\mathcal L_{\mathrm{sku}}
 +\beta\mathcal L_{\mathrm H}+\gamma\mathcal L_{\mathrm N}.
\end{equation}
Weight comparisons hold the optimizer and HCL aggregation fixed.

Training computes predictions and task masks, constructs structural and rejection-conditioned pairs under a fixed tree version, and updates shared parameters using the normalized losses. Serving retains the candidate encoder and predictive heads while omitting pair construction, contrastive ancestor lookup, and bucket aggregation. End-to-end latency still depends on the deployed encoder and feature pipeline.

\textbf{Computational cost.} With $B$ anchors, $M$ comparisons per anchor, dimension $d$, and tree height $h$, transformed similarities cost $O(BMd)$ beyond the predictive encoder. Walking parent paths for ancestor lookup adds up to $O(BMh)$ work. The rejected-SKU mean costs $O(|\operatorname{leaves}(b_r^-)|d)$ per uncached bucket and is recomputed from current embeddings each batch with stop-gradient. Table~\ref{tab:efficiency} reports training and serving costs separately.

\textbf{Gradient and sampling controls.} Comparisons are identified by event, anchor, positive, candidate set, and tree version. The ablations share pair construction, seed policy, and training budget. Each HCL anchor has 129 candidates (one positive and 128 negatives); similarity computation uses L2 normalization and $\varepsilon=10^{-8}$, with gradient norms capped at 5. Temperature $\tau$ scales logits and does not replace norm stabilization.

\section{Experiments}
The evaluation tests ranking gains beyond identical catalog features, sensitivity to hierarchy and rejection sampling, and online value alongside user experience.
\subsection{Evaluation Setup}
\subsubsection{Data and temporal protocol.}\label{sec:data}
The industrial data come from Fliggy's auxiliary-selection workflow, collected from June 1 to October 1, 2025. The test set covers October 1, 2025; the exact training and validation windows are June 1--September 28, 2025 and September 29--30, 2025. Historical features and catalog snapshots use information available before prediction time. The records cover 4.2 million users, 960 SKUs, and 6 xitems; repeated sessions follow one record per displayed SKU per session.

Table~\ref{tab:dataset} reports sample and event counts. Positives are clicked-and-booked products; negatives are the remaining sampled records. Click and purchase attribution windows are 30 minutes and 24 hours. Explicit replacement or removal after exposure yields 0.48 million rejection events across the splits. Negative-to-positive ratios are 10:1, 6.5:1, and 17:1 for train/validation/test. Unexposed offers are not treated as observed non-clicks.

\begin{table}[!htbp]
\centering
\caption{Dataset statistics (M: million). Positive samples denote clicked-and-booked products; task-event counts are reported separately.}
\label{tab:dataset}
\small\renewcommand{\arraystretch}{\papertablestretch}
\begin{tabular*}{\columnwidth}{@{\extracolsep{\fill}}lrrr@{}}
\toprule
\textbf{Records} & \textbf{Train} & \textbf{Valid.} & \textbf{Test} \\
\midrule
All samples & 11M & 1.5M & 1.8M \\
Positive & 1M & 0.2M & 0.1M \\
Negative & 10M & 1.3M & 1.7M \\
\midrule
Exposures & 11M & 1.5M & 1.8M \\
Clicks & 3.30M & 0.46M & 0.52M \\
Purchases & 1.00M & 0.20M & 0.10M \\
Rejections & 0.36M & 0.05M & 0.07M \\
\bottomrule
\end{tabular*}
\end{table}

\begin{table*}[!t]
\centering
\caption{Offline comparison. N@10/N@15 denote NDCG@10/NDCG@15. Learned-model rows report five-seed means; the bottom rows show variability for three key comparators. AUC is in percent and its SD in percentage points. $^\dagger$Price-rule rows use distinct exposure sets, unlike common-test-set learned models.}
\label{tab:offline}
\small\renewcommand{\arraystretch}{\papertablestretch}
\begin{tabular*}{\textwidth}{@{\extracolsep{\fill}}llrrr@{\hspace{1.5em}}llrrr@{}}
\toprule
\textbf{Group} & \textbf{Method} & \textbf{N@10} & \textbf{N@15} & \textbf{AUC} & \textbf{Group} & \textbf{Method} & \textbf{N@10} & \textbf{N@15} & \textbf{AUC} \\
\midrule
Rules & Cheapest$^\dagger$ & 0.8435 & 0.9146 & 77.15 & Multi-task & ESMM & 0.8801 & 0.9512 & 75.77 \\
 & Priciest$^\dagger$ & 0.8357 & 0.9120 & 73.73 & & MMoE & 0.8824 & 0.9536 & 76.12 \\
 & Policies & 0.8558 & 0.9231 & 77.17 & & PLE & 0.8845 & 0.9556 & 76.45 \\
\cmidrule(lr){1-5}\cmidrule(lr){6-10}
Predictive & LR & 0.8563 & 0.9301 & 75.71 & Hierarchy & HAKG & 0.8750 & 0.9421 & 76.20 \\
 & GBDT & 0.8661 & 0.9316 & 76.30 & Pre-training & SRP4CTR & 0.8790 & 0.9468 & 76.50 \\
 & WDL & 0.8697 & 0.9331 & 76.07 & Two-stage & TWIN & 0.8862 & 0.9545 & 76.95 \\
 & DIN & 0.8715 & 0.9380 & 76.15 & & TWIN V2 & 0.8900 & 0.9598 & 77.30 \\
 & ECAU-Net & 0.8732 & 0.9413 & 76.28 & Ours & PEETS & 0.8956 & 0.9667 & 78.20 \\
\midrule
\multicolumn{10}{@{}l@{}}{\footnotesize AUC (mean $\pm$ SD): Base $77.05\pm0.041$;\quad TWIN V2 $77.30\pm0.038$;\quad PEETS $78.20\pm0.045$.}\\
\multicolumn{10}{@{}l@{}}{\footnotesize N@10 (mean $\pm$ SD): Base $0.8774\pm0.0012$;\quad TWIN V2 $0.8900\pm0.0009$;\quad PEETS $0.8956\pm0.0010$.}\\
\bottomrule
\end{tabular*}
\end{table*}

\subsubsection{Baselines and implementation.}
The PEETS project URL is \url{https://github.com/BOTAK0803/PEETS}. The data are de-identified, and data use and online experimentation were internally authorized by the company. Raw industrial data cannot be publicly released.

The comparison set includes price-ordering rules and the production Policies control; LR~\cite{lavalley2008logistic}, GBDT~\cite{ke2017lightgbm}, WDL, DIN, and ECAU-Net~\cite{tao2023event}; ESMM, MMoE, and PLE; HAKG; SRP4CTR; and TWIN/TWIN V2. Baselines use public implementations adapted to the auxiliary SKU task. Cheapest and Priciest are evaluated on their respective policy-induced exposure sets, so their AUCs are contextual results rather than reversed rankings on one common test set.

Learned-model comparisons share the candidate pool, labels, temporal split, and available raw features, with at most 30 validation trials per method. Neural models use matched embedding dimensions and prediction-task definitions; these settings do not apply to rule or tree-based methods. Matched neural controls use 20 history events, while TWIN retains native retrieval. Price-rule exposure sets are excluded from claims about common-test-set performance.

\subsubsection{Metrics and optimization.}
We report AUC~\cite{fawcett2006roc} and request-level NDCG@10/@15~\cite{wang2013theoretical} using binary purchase labels and the probability score $p^{\mathrm{click}}p^{\mathrm{buy}\mid\mathrm{click}}$, without price weighting. Each request ranks its eligible displayed SKUs and may contain multiple purchased items. Let $n_q$ be its candidate count, $m_q$ its number of positives, and $\pi_q(j)$ the item at rank $j$. For requests $\mathcal Q_+$ with $m_q>0$,
\begin{equation}
\label{eq:ndcg}
 \mathrm{NDCG}@K=\frac{1}{|\mathcal Q_+|}
 \sum_{q\in\mathcal Q_+}
 \frac{\sum_{j=1}^{\min(K,n_q)}
 y_{q,\pi_q(j)}/\log_2(j+1)}
 {\sum_{j=1}^{\min(K,m_q)}1/\log_2(j+1)}.
\end{equation}
Requests without a positive item are excluded from NDCG but retained for AUC.

PEETS and the neural controls use Adam~\cite{kingma2014adam} with learning rate 0.0001, batch size 512, and 3 training epochs. PEETS uses dimension 128, temperature 0.07, and history limit 20. Learned-model results average 5 independent training seeds: 42, 1024, 2024, 2025, and 2026. Table~\ref{tab:offline} gives AUC and NDCG@10 standard deviations for Base, TWIN V2, and PEETS. Validation selects checkpoints by AUC, breaking ties by NDCG@10.

\textbf{Loss-weight selection.} Relative weights are $r=1$, $\beta=0.05$, and $\gamma=0.02$, with scale $c=1$. The validation search uses $r\in\{0.5,1,2\}$; $\beta,\gamma\in\{0.005,0.01,0.02,0.05,0.1\}$ and selection procedure 3 task-ratio trials followed by 25 auxiliary-weight trials, holding loss normalization, backbone, and training budget fixed. 

\subsection{Offline Evaluation}
\subsubsection{Overall comparison.}
Table~\ref{tab:offline} compares predictive-ranking performance. PEETS attains the highest AUC, NDCG@10, and NDCG@15 among the learned models evaluated on the common test set. Relative to TWIN V2, the strongest comparator on all three metrics, it improves AUC by 0.90 percentage points and NDCG@10 by 0.63\% relative. The respective NDCG@15 values are 0.9667 and 0.9598. The AUC gap exceeds the observed seed SDs of 0.038 and 0.045 points for TWIN V2 and PEETS. These SDs summarize training variation on the fixed test set. Matched-backbone controls further isolate the auxiliary objectives.

\begin{figure*}[!t]
\centering
\includegraphics[width=\textwidth]{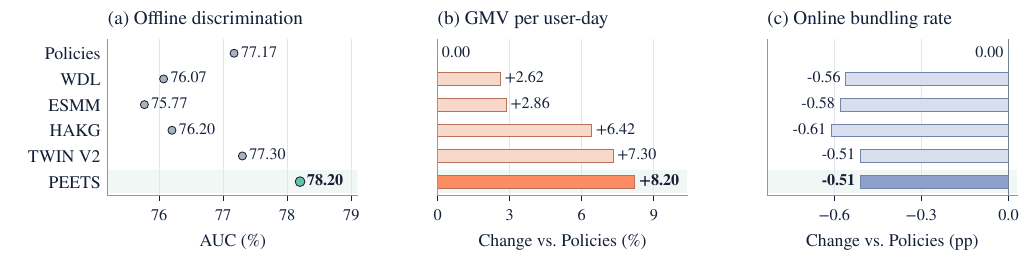}
\caption{Selected results from Tables~\ref{tab:offline} and~\ref{tab:online}. GMV per exposed user-day change is relative to Policies; BR change is in percentage points (pp). The panels show point estimates under their respective evaluation settings; overall GMV intervals appear in Section~\ref{sec:online}.}
\label{fig:results}
\figdesc{Three aligned panels compare offline AUC, online GMV per exposed user-day change, and online bundling-rate change for Policies, WDL, ESMM, HAKG, TWIN V2, and PEETS. PEETS has the highest AUC and GMV per exposed user-day, but its bundling rate is below Policies.}
\end{figure*}

Figure~\ref{fig:results} contrasts accuracy with commercial outcomes for comparators shared by the offline and online tables. PEETS has the highest AUC and GMV per exposed user-day among these rows, while its bundling rate is below Policies. Thus, value uplift should not be interpreted as an attachment-rate improvement.

\subsubsection{Matched-backbone ablation.}
Table~\ref{tab:ablation} compares bucket features, ordinary contrastive learning, HCL, and NCL on the same predictive backbone. PEETS improves AUC by 1.15 percentage points and NDCG@10 by 2.07\% relative to Base. Its AUC exceeds the bucket-feature and ordinary-CL controls by 0.88 and 0.74 points, respectively. Thus the observed benefit extends beyond adding catalog features. All ablations share data splits, raw inputs, catalog information, the MLP, training epochs, and seeds. Derived bucket features and auxiliary objectives are enabled only in the named variants. Ordinary CL retains the HCL/NCL pair samplers, loss reductions, and weights, replacing their transformed similarities with ordinary cosine similarity.

Both single-component variants improve all three metrics over Base. HCL reaches 77.89\% AUC and NCL reaches 77.72\%; combining them yields 78.20\%, a further 0.31 and 0.48 points, respectively. The combination has the highest means on all three metrics. Figure~\ref{fig:granularity} shows the same component AUCs at 15 buckets.

\begin{table}[!htbp]
\centering
\caption{Matched-backbone five-seed means. Top: feature and objective controls. Bottom: NCL component controls, retaining HCL and the original NCL sampling while changing only the shift.}
\label{tab:ablation}
\small\renewcommand{\arraystretch}{\papertablestretch}
\begin{tabular*}{\columnwidth}{@{\extracolsep{\fill}}lrrr@{}}
\toprule
\textbf{Variant} & \textbf{N@10} & \textbf{N@15} & \textbf{AUC (\%)} \\
\midrule
Base & 0.8774 & 0.9484 & 77.05 \\
Base + bucket features & 0.8821 & 0.9538 & 77.32 \\
Base + ordinary CL & 0.8845 & 0.9556 & 77.46 \\
Base + NCL & 0.8892 & 0.9601 & 77.72 \\
Base + HCL & 0.8918 & 0.9625 & 77.89 \\
PEETS & 0.8956 & 0.9667 & 78.20 \\
\midrule
\multicolumn{2}{@{}l}{\textbf{NCL component control}} & \textbf{N@10} & \textbf{AUC (\%)} \\
\multicolumn{2}{@{}l}{HCL + NCL, no shift} & 0.8929 & 77.98 \\
\multicolumn{2}{@{}l}{HCL + NCL, no mean} & 0.8927 & 77.97 \\
\bottomrule
\end{tabular*}
\end{table}

\subsubsection{Price-policy sensitivity.}
Table~\ref{tab:alpha} varies $\alpha$ over the same 14-day window with equal exposure across settings and a fixed predictive model. GMV increases across the listed $\alpha$ values, reaching CNY 173260 at $\alpha=6$, 22.25\% above $\alpha=0$. Over the same comparison, tickets rise by 0.79\% while BR falls by 0.91 percentage points. BR first declines and then recovers, so the trade-off is not monotonic in every outcome. Across this sweep, the highest GMV accompanies a lower BR than at $\alpha=0$.

\begin{table}[!htbp]
\centering
\caption{Price-policy sensitivity. Revenue is gross auxiliary GMV in CNY. The sweep varies the price exponent $\alpha$.}
\label{tab:alpha}
\small\renewcommand{\arraystretch}{\papertablestretch}
\begin{tabular*}{\columnwidth}{@{\extracolsep{\fill}}rrrr@{}}
\toprule
\textbf{$\alpha$} & \textbf{Tickets} & \textbf{BR (\%)} & \textbf{Revenue} \\
\midrule
0 & 8697 & 34.23 & 141725 \\
1 & 8690 & 34.11 & 143457 \\
2 & 8482 & 32.17 & 153671 \\
3 & 8560 & 32.55 & 160420 \\
4 & 8625 & 32.84 & 165800 \\
5 & 8698 & 33.12 & 170150 \\
6 & 8766 & 33.32 & 173260 \\
\bottomrule
\end{tabular*}
\end{table}

\subsubsection{Hierarchy and rejection sensitivity.}
\begin{figure}[!htbp]
\centering
\includegraphics[width=\linewidth]{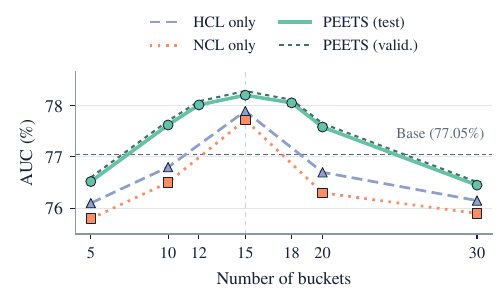}
\caption{Bucket-count sensitivity. PEETS includes validation and test AUC at seven settings; HCL/NCL are shown at five test settings. Base is a fixed reference. Lines connect the evaluated settings.}
\label{fig:granularity}
\figdesc{PEETS validation and test AUC across 5, 10, 12, 15, 18, 20, and 30 buckets, alongside HCL-only and NCL-only test curves at five settings and Base at 77.05 percent. PEETS test AUC is 78.20 at 15 buckets, 78.01 at 12, and 78.05 at 18; 5 and 30 buckets fall below Base.}
\end{figure}

Figure~\ref{fig:granularity} shows AUC across 5--30 buckets. PEETS peaks at 15 buckets, selected using validation AUC (78.28\%) before final test evaluation (78.20\%). Nearby settings at 12 and 18 buckets reach 78.01\% and 78.05\% test AUC. The 10/20-bucket results are 77.62\%/77.58\%. At 5 and 30 buckets, AUC falls to 76.52\% and 76.45\%, below Base's 77.05\%. The tested 12--18-bucket settings remain above Base, whereas 5 and 30 buckets fall below it.

For 15 buckets, multiplying each internal quantile probability by an independent $U(0.95,1.05)$ draw, sorting, and rebuilding boundaries yields 78.16\% validation and 78.08\% test AUC. Equal-width bins yield 78.10\%/78.02\%. These two settings lower test AUC by 0.12 and 0.18 points relative to the default quantile partition.

The additional controls in Table~\ref{tab:ablation} fix HCL, rejection events, anchor--positive--negative pairs, reductions, weights, and seeds. Removing the SKU mean gives $\Delta_r=e_{b_r^-}-e_{b_r^{\mathrm{ref}}}$ and 77.97\% AUC; setting $\Delta_r=0$ gives 77.98\%. Full PEETS exceeds them by 0.23 and 0.22 points, with NDCG@10 gains of 0.0029 and 0.0027. No shift retains the NCL loss and its rejection-conditioned sampling; unlike ordinary CL, it preserves transformed HCL. These comparisons report differences between the five-seed means.

\subsection{Targeted Revenue Analysis}
Table~\ref{tab:analysis} reports PEETS's relative improvement in GMV per request over TWIN V2 on selected online subsets. For each arm and subset, GMV is divided by its request count before computing the relative change. The Requests column totals requests from both arms. The ambiguous-candidate subset is defined by requests containing at least 3 SKUs of one xitem using pre-outcome information. Low- and high-frequency groups use the prior-activity threshold at most 2 versus more than 2 interactions in 30 days. The 95\% intervals use 1000 bootstrap resamples of randomized users and the percentile method, keeping each user's requests together.

\begin{table}[!htbp]
\centering
\caption{Selected online subsets: relative change in GMV per request over TWIN V2, with user-bootstrap 95\% intervals. Requests are pooled across the two arms.}
\label{tab:analysis}
\small\renewcommand{\arraystretch}{\papertablestretch}
\begin{tabular*}{\columnwidth}{@{\extracolsep{\fill}}lrrr@{}}
\toprule
\textbf{Subset} & \textbf{Requests} & \textbf{$\Delta$ (\%)} & \textbf{95\% CI} \\
\midrule
Ambiguous & 350000 & +3.6 & [3.12,4.08] \\
Low-frequency & 420000 & +4.2 & [3.65,4.75] \\
High-frequency & 980000 & +1.1 & [0.82,1.38] \\
\bottomrule
\end{tabular*}
\end{table}

All three subset intervals lie above zero. Low- and high-frequency groups total 1.4 million filtered requests. Ambiguity is defined separately from activity, so its request count is reported separately. Subset results use GMV per request, while the overall comparison uses GMV per exposed user-day.

\subsection{Online Evaluation}\label{sec:online}
The online experiment spans 14 days, with user-level randomized assignment to seven arms. Mean daily exposed-user counts range from 20371 to 20442 across arms. The experiment uses 10\% of traffic from September 1 to September 14, 2025. Exposure is counted per user-day; Table~\ref{tab:online} reports the daily mean. Policies is the production reference. The initial checkpoint uses data through August 31, 2025, with daily updates using records available by the preceding day. The October offline retrospective evaluation uses separately trained models.

Revenue is gross auxiliary GMV in CNY before refunds. Table~\ref{tab:online} gives daily means over 14 days. GMV per exposed user-day is mean daily GMV divided by mean daily exposed users. Because exposures differ across arms, its relative change differs from that of daily total GMV. Users may recur across days. Sample-ratio and cross-arm contamination checks found no anomalies. 

\begin{table}[!htbp]
\centering
\caption{Online results: per-arm daily means over 14 days. Revenue is gross auxiliary GMV in CNY before refunds.}
\label{tab:online}
\small\renewcommand{\arraystretch}{\papertablestretch}
\begin{tabular*}{\columnwidth}{@{\extracolsep{\fill}}lrrrr@{}}
\toprule
\textbf{Arm} & \textbf{Exposed} & \textbf{Tickets} & \textbf{BR (\%)} & \textbf{Revenue} \\
\midrule
Policies & 20382 & 8561 & 33.83 & 159764 \\
WDL & 20435 & 8602 & 33.27 & 164378 \\
ESMM & 20419 & 8627 & 33.25 & 164635 \\
HAKG & 20371 & 8611 & 33.22 & 169922 \\
TWIN V2 & 20442 & 8671 & 33.32 & 171927 \\
Base & 20395 & 8640 & 33.28 & 167850 \\
PEETS & 20428 & 8766 & 33.32 & 173260 \\
\bottomrule
\end{tabular*}
\end{table}

PEETS improves GMV per exposed user-day by 8.20\% over Policies (95\% CI [6.84, 10.12]), 3.06\% over Base ([1.85, 4.61]), and 0.84\% over TWIN V2 ([0.16, 1.41]). Each interval uses 1000 within-arm user bootstrap resamples, retaining all 14 days per sampled user and recomputing total GMV divided by exposed user-days before the relative comparison. Percentile limits are 2.5\% and 97.5\%. All three pointwise intervals exclude zero. Unnormalized daily GMV increases by 8.45\%, 3.22\%, and 0.78\%, respectively. Relative to Policies, tickets rise by 2.39\% while BR falls by 0.51 points. The absolute daily GMV difference is CNY 13496 per arm in the measured traffic.

\textbf{User-experience guardrails.} Table~\ref{tab:guardrails} compares PEETS with Policies: auxiliary-tagged complaints per exposed user, all-cause cancellations per booked primary ticket, and abandonment per initiated checkout. Post-booking outcomes use seven-day follow-up. Differences are treatment minus control, so positive values indicate deterioration. The 95\% intervals use 1000 user-level bootstrap resamples and the percentile method. The non-inferiority criterion is an upper confidence limit below the acceptable increase specified before the experiment.

\begin{table}[!htbp]
\centering
\caption{User-experience guardrails. Rates are in percent; treatment-minus-control differences, intervals, and margins are in percentage points.}
\label{tab:guardrails}
\footnotesize\setlength{\tabcolsep}{2pt}\renewcommand{\arraystretch}{\papertablestretch}
\begin{tabular*}{\columnwidth}{@{\extracolsep{\fill}}lrrrrr@{}}
\toprule
\textbf{Metric} & \textbf{Ctrl.} & \textbf{PEETS} & \textbf{$\Delta$} & \textbf{95\% CI ($\Delta$)} & \textbf{Margin} \\
\midrule
Complaints & 0.12 & 0.13 & +0.01 & [-0.009,0.028] & 0.03 \\
Cancellations & 1.80 & 1.83 & +0.03 & [-0.076,0.137] & 0.10 \\
Abandonment & 18.60 & 18.55 & -0.05 & [-0.251,0.152] & 0.20 \\
\bottomrule
\end{tabular*}
\end{table}

The upper limits for complaints (0.028) and abandonment (0.152) are below their respective margins of 0.03 and 0.20 points. The cancellation interval's upper limit is 0.137 points, above the 0.10-point margin; the prespecified cancellation non-inferiority criterion is not met.

\section{Session-level Diagnostic}\label{sec:case}
We retrospectively score a logged Beijing--Singapore session with both models; only the served policy's feedback is observed. The fixed insurance candidates A--D cost CNY 30, 60, 68, and 25: basic accident, premium accident plus delay, comprehensive travel, and delay-only coverage. The user removed A and purchased B.

For A--D, the probability scores are 0.25, 0.15, 0.12, 0.28 (Base) and 0.04, 0.34, 0.31, 0.11 (PEETS). Equation~\ref{eq:score} with the fixed $\alpha=6$ yields Base's order B$>$C$>$A$>$D and PEETS's C$>$B$>$D$>$A. In this replay, B moves from third to first in probability rank and from first to second in price-weighted rank.

\section{Deployment and Practical Considerations}
PEETS trains on Alibaba's AI Platform (PAI) and serves through Real-Time Prediction (RTP) and Personalization Platform (TPP). Table~\ref{tab:efficiency} reports training time and server-side latency from request receipt to response, including feature processing and model computation. The configuration uses 8 A10 training workers and 16-vCPU serving instances with FP32 and batch size 1.

\begin{table}[!htbp]
\centering
\caption{Training and serving efficiency on 11M records. Neural models train for 3 epochs; latencies are in ms.}
\label{tab:efficiency}
\small\renewcommand{\arraystretch}{\papertablestretch}
\begin{tabular*}{\columnwidth}{@{\extracolsep{\fill}}lrrrr@{}}
\toprule
\textbf{Method} & \textbf{Train time} & \textbf{Mean} & \textbf{p50} & \textbf{p95} \\
\midrule
Policies & -- & 2 & 1.8 & 3.2 \\
GBDT & 34m\,19s & 14 & 13.2 & 21.6 \\
WDL & 58m\,28s & 27 & 26.1 & 42.8 \\
Base & 67m\,50s & 26 & 25.4 & 42.0 \\
PLE & 73m\,28s & 27 & 26.5 & 43.7 \\
TWIN V2 & 77m\,33s & 26 & 25.5 & 42.3 \\
PEETS & 73m\,30s & 26 & 25.5 & 42.1 \\
\bottomrule
\end{tabular*}
\end{table}

\textbf{Catalog versions.} Model checkpoints record their tree versions. Lookup failure uses the previous valid model and catalog snapshot.

\textbf{Cost and serving audit.} PEETS adds 5m\,40s (8.35\%) to Base's training time. Both record 26~ms mean latency, while PEETS's p50 and p95 are 0.1~ms higher. Latency is measured over the same 14-day window at 300 requests/s per serving instance.

\Needspace{6\baselineskip}
\section{Conclusion}
PEETS combines ancestor-conditioned and rejection-conditioned training with the existing serving path. Five-seed results and controlled NCL comparisons support the observed ranking gains; the tested 12/18-bucket and alternative partitions remain within 0.19 points of the default test AUC. GMV per exposed user-day improves by 8.20\%, 3.06\%, and 0.84\% over Policies, Base, and TWIN V2, with positive user-bootstrap intervals. PEETS records 42.1~ms p95 latency under the measured load. Its BR is below Policies, and the cancellation interval's upper limit exceeds the prespecified margin.

\bibliographystyle{IEEEtran}
\bibliography{references}
\end{document}